\pdfoutput=1
\documentclass[10pt, conference, compsocconf]{IEEEtran}
\usepackage{multirow}
\usepackage{amssymb}
\usepackage{textcomp} 
\usepackage{cite}

\ifCLASSINFOpdf
  \usepackage[pdftex]{graphicx}
\else
\fi
\usepackage[cmex10]{amsmath}
\usepackage[tight,footnotesize]{subfigure}
\usepackage{url}

\IEEEoverridecommandlockouts
\begin{document}
\setlength{\parskip}{0pt}
%
\title{Multi-Path Region-Based Convolutional Neural Network for Accurate Detection of Unconstrained ``Hard Faces''}


\author{\IEEEauthorblockN{Yuguang Liu, Martin D. Levine}
\IEEEauthorblockA{Department of Electrical and Computer Engineering\\
Center for Intelligent Machines, McGill University\\
Montreal, QC., Canada\\
yuguang@cim.mcgill.ca, levine@cim.mcgill.ca}
}
%
\maketitle

\begin{abstract}
Large-scale variations still pose a challenge in unconstrained face detection. To the best of our knowledge, no current face detection algorithm can detect a face as large as $800 \times 800$ pixels while simultaneously detecting another one as small as $8 \times 8$ pixels within a single image with equally high accuracy. We propose a two-stage cascaded face detection framework, Multi-Path Region-based Convolutional Neural Network (MP-RCNN), that seamlessly combines a deep neural network with a classic learning strategy, to tackle this challenge. The first stage is a Multi-Path Region Proposal Network (MP-RPN) that proposes faces at three different scales. It simultaneously utilizes three parallel outputs of the convolutional feature maps to predict multi-scale candidate face regions. The ``atrous'' convolution trick (convolution with up-sampled filters) and a newly proposed sampling layer for ``hard'' examples are embedded in MP-RPN to further boost its performance. The second stage is a Boosted Forests classifier, which utilizes deep facial features pooled from inside the candidate face regions as well as deep contextual features pooled from a larger region surrounding the candidate face regions. This step is included to further remove hard negative samples. Experiments show that this approach achieves state-of-the-art face detection performance on the WIDER FACE dataset ``hard'' partition, outperforming the former best result by 9.6\% for the Average Precision.

\end{abstract}


%
\IEEEpeerreviewmaketitle

\section{Introduction}
Although face detection has been extensively studied during the past two decades, detecting unconstrained faces in images and videos has not yet been convincingly solved. Most classic and recent deep learning methods tend to detect faces where fine-grained facial parts are clearly visible. This negatively affects their detection performance in the case of faces at low-resolution or out-of-focus blur, which are common issues in surveillance camera data. The lack of progress in this regard is largely due to the fact that current face detection benchmark datasets (e.g., FDDB~\cite{r01}, PACAL FACE~\cite{r02} and AFW~\cite{r03}) are biased towards high-resolution face images with limited variations in scale, pose, occlusion, illumination, out-of-focus blur and background clutter. Recently, a new face detection benchmark dataset, WIDER FACE~\cite{r04}, has been released to tackle this problem. WIDER FACE consists of 32,203 images with 393,703 labeled faces. Images in WIDER FACE also have the highest degree of variations in scale, pose, occlusion, lighting conditions, and image blur. As indicated in the WIDER FACE report~\cite{r04}, of all the factors that affect face detection performance, scale is the most significant. 

In view of the challenge created by facial scale variation in face detection, we propose a Multi-Path Region-based Convolutional Neural Network (MP-RCNN) to detect big faces and tiny faces with high accuracy. At the same time, it is noteworthy that by virtue of the abundant feature representation power of deep neural networks and the employment of contextual information, our method also possesses a high level of robustness to other factors. These are a consequence of variations in pose, occlusion, illumination, out-of-focus blur and background clutter, as shown in Figure 1.

\begin{figure*}
\begin{center}
\includegraphics[width=0.8\linewidth]{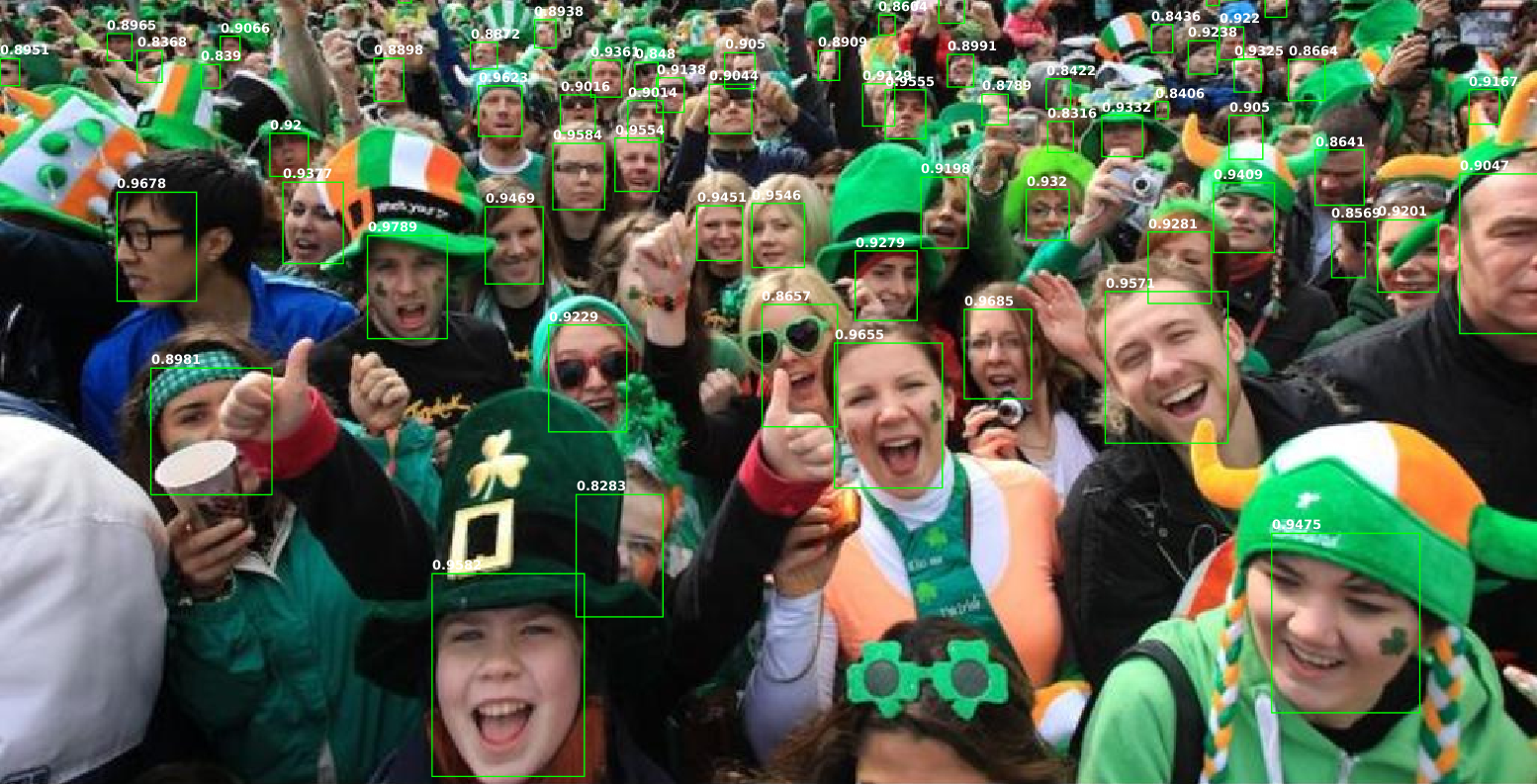}
\end{center}
   \caption{An example of face detection results on the WIDER FACE dataset~\cite{r01} using the proposed MP-RCNN method. We observe that it can robustly detect unconstrained ``hard faces'' with large variations in scale, pose, occlusion, lighting conditions, and image blur.}
\label{fig:short}
\end{figure*}

MP-RCNN is composed of two stages. The first stage is a Multi-Path Region Proposal Network (MP-RPN) that proposes faces at three different scales: small (8-32 pixels in height), medium (32-360 pixels in height) and large (360-900 pixels in height). These scales cover the majority of faces available in all public face detection databases, e.g., WIDER FACE~\cite{r04}, FDDB~\cite{r01}, PASCAL FACE~\cite{r02} and AFW~\cite{r03}. We observe that the feature maps of lower-level convolutional layers are most sensitive to small-scale face patterns, but almost agnostic to large-scale face patterns due to a limited receptive field. Conversely, the feature maps of the higher-level convolutional layers respond strongly to large-scale face patterns while ignoring small-scale patterns. On the basis of this observation, we simultaneously utilize three parallel outputs of the convolutional feature maps to predict multi-scale candidate face regions. We note that the path of medium-scale (32-360) and large-scale (360-900) span a much larger scale range than the small-scale (8-32) path does. Thus we additionally employ the so-called ``atrous'' convolution trick (convolution with up-sampled filters)~\cite{r05} together with normal convolution to acquire a larger field of view so as to comprehensively cover the particular face scale range. Moreover, a newly proposed sampling layer is embedded in MP-RPN to further boost the discriminative power of the network for difficult face/non-face patterns. 

To further contend with difficult false positives while including difficult false negatives, we add a second stage Boosted Forests classifier after MP-RPN. The Boosted Forests classifier utilizes deep facial features pooled from inside the candidate face regions. It also invokes deep contextual features pooled from a larger region surrounding candidate face regions to make a more precise prediction of face/non-face patterns.

Our MP-RCNN achieves state-of-the-art detection performance on both the WIDER FACE~\cite{r04} and FDDB ~\cite{r01} datasets. In particular, on the most challenging so-called ``hard'' partition of the WIDER FACE test set that contains just small faces, we \textit{outperform} the former best result by 9.6\% for the Average Precision.

The rest of the paper is organized as follows. Section 2 reviews related work. Section 3 introduces the proposed MP-RCNN approach to the problem of unconstrained face detection. Section 4 presents experimental results to demonstrate the rationale behind our network design and compares our method with other state-of-the-art face detection algorithms on the WIDER FACE~\cite{r04} and FDDB~\cite{r01} datasets. Section 5 concludes the paper and proposes future work.

\section{Related work}
There are two established sets of methods for face detection, one based on deformable part models ~\cite{r02,r03} and the other on rigid templates~\cite{r06,r07,r08,r09}. Prior to the resurgence of Convolutional Neural Networks (CNN)~\cite{r10}, both sets of methods relied on a combination of ``hand-crafted'' feature extractors to select facial features and classic learning methods to perform binary feature classification. Admittedly, the performance of these face detectors has been increasingly improved by the use of more complex features~\cite{r07,r08,r11} or better training strategies~\cite{r03,r06,r12}. Nevertheless, using ``hand-crafted'' features and classic classifiers has stymied the development of seamlessly connecting feature selection and classification in a single computational process. In general, they require that many hyper-parameters be heuristically set. For example, both~\cite{r12} and~\cite{r11} needed to divide the training data into several partitions according to face poses and train a separate model for each partition.

Deep neural networks, with its seamless concatenation of feature representation and pattern classification, have become the current trend of rigid templates for face detection. Farfade et al.~\cite{r13} proposed a single Convolutional Neural Network (CNN) model based on AlexNet~\cite{r10} to deal with multi-view face detection. Li et al.~\cite{r14} used a cascade of six CNNs for alternative face detection and face bounding box calibration. However, these two methods need to crop face regions and rescale them to specific sizes. This increases the complexity of the training and testing. Thus they are not suitable for efficient unconstrained face detection where faces of different scales coexist in the same image. Yang et al.~\cite{r15} proposed applying five parallel CNNs to predict five different facial parts, and then evaluate the degree of face likeliness by analyzing the spatial arrangement of facial part responses. The usage of facial parts makes the face detector more robust to partial occlusions, but like DPM based face detectors, this method can only deal with faces of relatively large size.

Recently, Faster R-CNN~\cite{r16}, a deep learning framework, achieved state-of-the-art object detection  because of two novel components. The first is a Region Proposal Network (RPN) to recommend object candidates of different scales and aspect ratios. The second is a Region-based Convolutional Neural Network (RCNN) to pool the object candidates to construct a fixed-length feature vector, which is employed to make a prediction. Zhu et al.~\cite{r17} proposed a Contextual Multi-Scale Region-based CNN (CMS-RCNN) face detector, which extended Faster RCNN~\cite{r16} in two respects. First, RPN was replaced by a Multi-Scale Region Proposal Network (MS-RPN) to propose face regions based on the combined information from multiple convolutional layers. Secondly, a Contextual Multi-Scale Convolution Neural Network (CMS-CNN) was proposed to replace RCNN for pooling features. This was not restricted to the last convolutional layer, as in RCNN, but also from several lower level convolutional layers. In addition, contextual information was also pooled to promote robustness. Thus MS-RCNN~\cite{r17} has indeed improved RPN by combining feature maps from multiple convolutional layers in order to make a proposal. However, it is necessary to down-sample the lower-level feature maps to concatenate the feature maps of the last convolutional layer. This down-sampling design inevitably diminishes the network's discriminative power for small-scale face patterns. 

The Multi-Path Region Proposal Network (MP-RPN) presented in this paper enhances the discriminative power by eliminating the down-sampling and concatenation steps and directly utilizes feature maps at different resolutions. It proposes faces at different scales: lower-level feature maps are used to propose small-scale faces, while higher-level feature maps do so for large-scale faces. In this way, the scale-aware discriminative power of different feature maps is fully exploited. 

It has been pointed out~\cite{r18} that the Region-of-Interest (ROI) pooling layer applied to low-resolution feature maps can lead to ``plain'' features due to the bins collapsing. We note that this “lost” information will lead to non-discriminative small regions. However, since detecting small-scale faces is one of the main objectives of this paper, we have instead pooled features from lower-level feature maps to reduce information collapsing. For example, we reduce information collapsing by using conv3\_3 and conv4\_3 of VGG16~\cite{r19}, which have higher resolution, instead of conv5\_3 of VGG16~\cite{r19} used by Faster RCNN~\cite{r16} and CMS-RCNN~\cite{r17}. The pooled features are then trained by a Boosted Forest (BF) classifier as is done for pedestrian detection~\cite{r18}. But unlike~\cite{r18}, we also pool contextual information in addition to the facial features to further boost detection performance. 

Although the practice of adding a BF classifier makes our method not an end-to-end deep neural network solution, the combination of MP-RPN and a BF classifier has two advantages. First, features pooled from different convolutional layers need not be normalized before concatenation since the BF classifier treats each element of a feature vector separately. In contrast, in CMS-RCNN~\cite{r17}, three different normalization scales need to be carefully selected to concatenate the RoI features from three convolutional layers. Secondly, both MP-RPN and the BF classifier only need to be trained once, which is as efficient as the four-step alternative training process used in Faster RCNN~\cite{r16} and CMS-RCNN~\cite{r17}.

The proposed MP-RPN shares some similarity with the Single Shot Multibox Detector (SSD)~\cite{r20} and the Multi-Scale Convolutional Neural Network (MS-CNN)~\cite{r21}. Both methods use multi-scale feature maps to predict objects of different sizes in parallel. However, our work differs from these in two notable respects. First, we employ a fine-grained path to classify and localize tiny faces (as small as $8\times 8$ pixels). Both SSD and MS-CNN lack such a characteristic since both were proposed to detect general objects, such as cars or tables, which have a much larger minimum size. Second, for medium- and large-scale path, we additionally employ the ``atrous'' convolution trick (convolution with up-sampled filters)~\cite{r05} together with the normal convolution to acquire a larger field of view. In this way, we are able to use three paths to cover a large spectrum of face sizes, from $8\times 8$ to $900\times 900$ pixels. By comparison, SSD~\cite{r20} utilized six paths to cover different object scales, which makes the network much more complex.

\section{Approach}
In this section, we introduce the proposed MP-RCNN face detector, which consists of two stages: a Multi-Path Region Proposal Network (MP-RPN) for the generation of face proposals and a Boosted Forest (BF) for the verification of face proposals.

\subsection{Multi-Path Region Proposal Network}
The detailed architecture of a Multi-Path Region Proposal Network (MP-RPN) is shown in Figure 2. Given a full image of arbitrary size, MP-RPN proposes faces through three detection branches: Det-4 for proposing small-scale faces (8-32 pixels in height), Det-16 for medium-scale faces (32-360 pixels in height) and Det-32 for large-scale faces (360-900 pixels in height). We adopt the VGG-16 net~\cite{r19} (from Conv1\_1 to Conv5\_3) as the CNN trunk and the three detection branches emanate from different layers of the trunk. Since the branches of Det-4 and Det-16 stay close to the lower layers of the trunk network, they affect the gradients of the corresponding lower layers more than the Det-32 branch. Thus we add L2 normalization layers~\cite{r22} to these two branches to avoid the potential learning instability.

\begin{figure*}
\begin{center}
\includegraphics[width=0.95\linewidth]{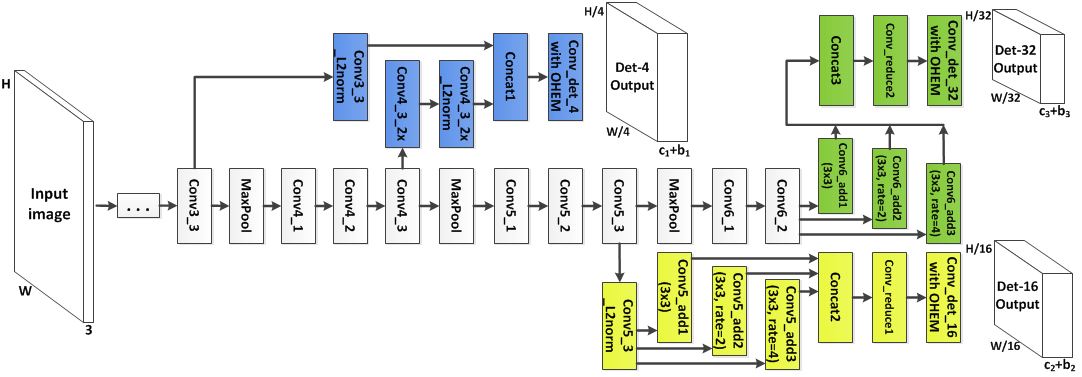}
\end{center}
   \caption{The architecture of MP-RPN. Three detection paths branch from the CNN trunk: Det-4 (in blue), Det-16 (in yellow) and Det-32 (in green). The bold cubes are the output tensors of the network. $c_i$ is the number of anchors, and $b_i$ is the number of bounding box coordinates of a particular detection branch ($i=1$ for Det-4, 2 for Det-16, and 3 for Det-32).}
\label{fig2}
\end{figure*}

Similar to RPN in Faster RCNN~\cite{r16}, for each detection branch, we slide a $3 \times 3$ convolutional network (Conv\_det\_4, Conv\_det\_16, and Conv\_det\_32 in Figure~\ref{fig2}) over the feature map of the prior convolutional layer (Concat1, conv\_reduce1, and conv\_reduce2 in Figure~\ref{fig2}). This convolutional layer is fully connected to a $3 \times 3$ spatial window of the input feature map. Each sliding window is mapped to a 512-dimensional vector. The vector is fed into two sibling fully connected layers, a box-classification layer ($c_i$ in Figure~\ref{fig2}, $i=1$ for Det-4 branch, $2$ for Det-16 branch, and $3$ for Det-32 branch) and a box-regression layer ($b_i$ in Figure~\ref{fig2}, $i=1$ for Det-4 branch, $2$ for Det-16 branch, and $3$ for Det-32 branch). At each sliding window location, we simultaneously predict $k$ region proposals of different scales (aspect ratio is always set to $1$). The $k$ proposals are parameterized relative to $k$ reference boxes, called anchors~\cite{r16}. Each anchor is centered at the sliding window and associated with a scale. The anchors are necessary because they refer to both the scale and position information so that face of different sizes located in any position of an image can be detected by the convolutional network. Table \ref{table0} shows the anchor scales (in pixel) allocated to each branch.
\begin{table}[h!]
\begin{center}
\caption{Anchor scales (in pixel) of each detection branch}
\label{table0}
    \begin{tabular}{  p{1.2cm} | p{1.5cm} | p{1.5cm} | p{1.5cm}}
    \hline
    Branch & Det-4 & Det-16 & Det-32 \\ \hline
    Anchor Scales & $8^2$, $16^2$, $32^2$ & $32^2$, $64^2$, $128^2$, $256^2$, $360^2$ & $360^2$, $512^2$, $720^2$, $900^2$ \\ 
    \hline
    \end{tabular}
\end{center}
\end{table}

During training, the parameters $W$ of the MP-RPN are learned from a set of $N$ training samples $S = \{ ({X_i},{Y_i})\} _{i = 1}^N$, where $X_i$ is an image patch associated with an anchor, and $Y_i=(p_i,b_i)$ the combination of its ground truth label $p_i=\{0,1\}$ ($0$ for non-face and $1$ for face) and ground truth box regression target ${b_i} = (b_i^x,b_i^y,b_i^w,b_i^h)$ associated with an ground truth face region. They are the parameterizations of the four coordinates following~\cite{r16}: 
$b_i^x = ({x_{gt}} - {x_i})/{w_i}, b_i^y = ({y_{gt}} - {y_i})/{h_i}, b_i^w = \log ({w_{gt}}/{w_i}), b_i^h = \log ({h_{gt}}/{h_i})$, where $x,y,w,h$ denote the two coordinates of the box center, width, and height. Variables $x_i,x_{gt}$ are for the image patch $X_i$ and its ground truth face region $X_i^{gt}$ respectively (likewise for $y$, $w$, and $h$).

We define the loss function for MP-RPN as
\begin{equation}
l(W) = \sum\limits_{m = 1}^M {{\alpha _m}} {L^m}({\{ ({X_i},{Y_i})\} _{i \in {S^m}}}|W)
\end{equation}
where $M=3$ is the number of detection branches, ${\alpha _m}$ is the weight of loss function $L^m$, and $S = \{ {S^1},{S^2},...,{S^M}\}$, where $S^m$ contains the training samples of the $m^{th}$ detection branch. The loss function for each detection branch contains two objectives
\begin{equation}
\begin{split}
{L^m}({\{ ({X_i},{Y_i})\} _{i \in {S^m}}}|W) &= \frac{1}{{{N_m}}}\sum\limits_{i \in {S^m}} {{L_{cls}}} (p({X_i}),{p_i}) \\&+ \lambda \left[\kern-0.15em\left[ {{p_i} = 1} 
 \right]\kern-0.15em\right]{L_{reg}}(b({X_i}),{b_i})
\end{split}
\end{equation}
where $N_m$ is the number of samples in the mini-batch of the $m^{th}$ detection branch, $p({X_i}) = ({p_0}({X_i}),{p_1}({X_i}))$ is the probability distribution over the two classes, non-face and face, respectively. $L_{cls}$ is the cross entropy loss, $b({X_i}) = ({b^x}({X_i}),{b^y}({X_i}),{b^w}({X_i}),{b^h}({X_i}))$ is the predicted bounding box regression target, $L_{reg}$ is the smoothL1 loss function defined in~\cite{r23} for bounding box regression and $\lambda$ is a trade-off coefficient between classification and regression. Note that $L_{reg}$ is computed only when a training sample is positive ($\left[\kern-0.15em\left[ {{p_i} = 1}\right]\kern-0.15em\right]$).

\subsubsection{Details of Each Detection Branch}
\underline{\textbf{\textit{Det-4}}}: Although Conv4\_3 layer (stride = 8 pixels) might seem to already be sufficiently discriminative on regions as small as $8\times 8$ pixels, this is not the case. We found in preliminary experiments that when a $8\times 8$ face happened to be located between two neighboring anchors, neither could be precisely regressed to the face location. Thus, to boost the localization accuracy of small faces, we instead use Conv3\_3 layer (with stride = 4 pixels) to propose small faces. At the same time, the feature maps of Conv4\_3 layer are up-sampled (by a deconvolution layer) and then concatenated to those of the Conv3\_3 layer. The higher-level Conv4\_3 layer provides Conv3\_3 layer with some ``contextual'' information and helps it to remove hard false positives.

\underline{\textbf{\textit{Det-16}}}: This detection branch is forked from Conv5\_3 layer to detect faces from $32\times 32$ to $360\times 360$ pixels. However, this large span of scales cannot be well accounted for by a single convolutional path. Inspired by the ``atrous'' spatial pyramid pooling~\cite{r05} used in semantic image segmentation, we employ three parallel convolutional paths: a normal $3\times 3$ convolutional layer, an ``atrous'' convolutional layer with ``atrous'' rate 2 and an ``atrous'' convolutional layer with ``atrous'' rate 4. These three convolutional layers have increasing receptive field sizes and are able to comprehensively cover the large face scale range.

\underline{\textbf{\textit{Det-32}}}: This detection branch is forked from Conv6\_2 layer to detect faces from $360\times 360$ to $900\times 900$ pixels. Similar to \underline{\textbf{\textit{Det-16}}}, three parallel convolutional paths are employed to fully cover the scale range.

\begin{figure*}
\begin{center}
\includegraphics[width=0.85\linewidth]{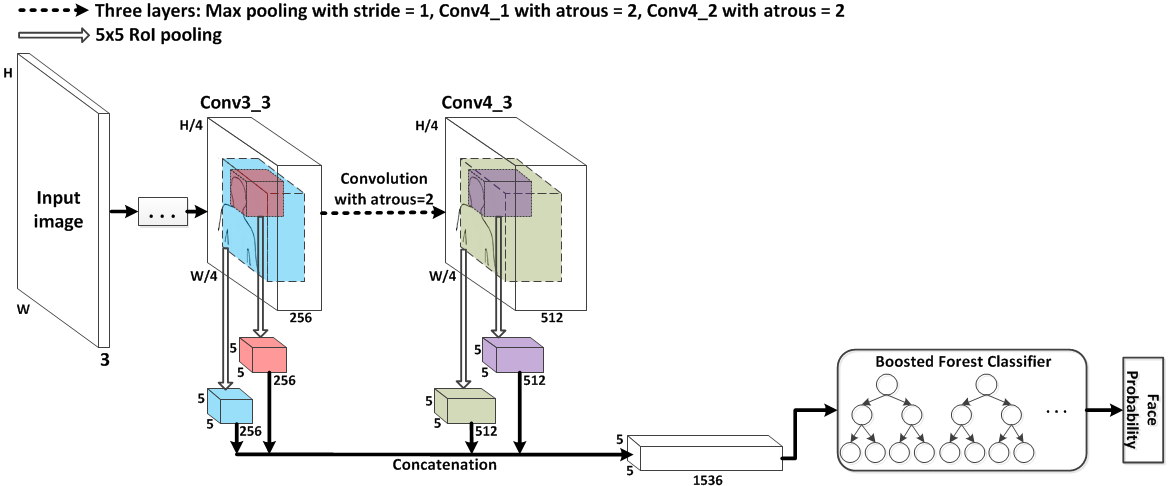}
\end{center}
   \caption{The architecture of the $2^{nd}$ stage. Features are pooled from both the proposal region (given by MP-RPN) and a larger ``contextual'' region in the feature maps of Conv3\_3 and Conv3\_4, then concatenated and finally fed into a boosted forest classifier for classification.}
\label{fig3}
\end{figure*}

\subsubsection{Online Hard Example Mining (OHEM) layer}
The training samples for MP-RPN are usually extremely unbalanced. This is because face regions are scarce compared to background (non-face) regions, so only a few anchors can be positive (matched to face regions) and most of the anchors are negative (matched to background regions). As indicated by~\cite{r24}, explicitly mining hard negative examples with high training loss leads to better training and testing performance than randomly sampling all negative examples. In this paper, we propose an Online Hard Example Mining (OHEM) layer specifically for MP-RPN. It is applied independently to each detection branch in Figure 2 in order to mine both hard positive and negative examples at the same time. We fix the selection ratio of hard positive examples and negative examples to 1:3, which experimentally provides more stable training. These selected hard examples are then used in back-propagation for updating network weights. 

Two steps are involved in the OHEM layer.
\underline{\textbf{\textit{Step 1}}}: Given all anchors (training samples) and their classification loss, we compare each anchor with its eight spatial neighbors (top, left, right, bottom, top-left, top-right, bottom-left and bottom-right). If the loss is greater than all of its neighbors, this anchor is kept as is; otherwise it is suppressed by setting its classification loss to zero.
\underline{\textbf{\textit{Step 2}}}: All anchors are sorted in the descending order of their classification loss and hard positive and negative samples are selected according to this order. The ratio between the selected positives and negatives was chosen as 1:3.

The proposed OHEM layer is ``online'' in the sense that it is seamlessly integrated into the forward pass of the network to generate a mini-batch of hard examples. Thus we do not need to freeze the training model to mine hard examples from all training data, and used the hard examples to update the current model.

Note that unlike~\cite{r24}, which proposed an OHEM layer for fast RCNN~\cite{r23}, here the OHEM layer is used in MP-RPN but it can also be generally used in other Region-based Proposal Networks, such as RPN in faster RCNN~\cite{r16} and MS-RPN in CMS-RCNN~\cite{r17}.

\subsection{Feature Extraction and Boosted Forest}
The detailed architecture of Stage 2 is shown in Figure~\ref{fig3}. Given a complete image of arbitrary size and a set of proposals provided by the MP-RPN, RoI pooling~\cite{r23} is used to extract features in the proposed regions from the feature maps of both Conv3\_3 and Conv4\_3. Conv3\_3 contains fine-grained information, while Conv4\_3, with a larger receptive field, implicitly contains “contextual” information. Similar to~\cite{r18}, the ``atrous'' convolution trick is employed to Conv4\_1, Conv4\_2 and Conv4\_3. This increases the resolution of the feature maps of Conv4\_3 to twice its original value. This change produces better experimental results.

Inspired by~\cite{r02, r17}, apart from extracting features from a proposed region, we also explicitly extract ``contextual'' features from a large region surrounding the proposal region. Suppose the original region is $[l, t, w, h]$, where $l$ is the horizontal coordinate of its left edge, $t$ the vertical coordinate of the top edge, and $w$, $h$ the width and height of the region, respectively. We set the corresponding “contextual” region to $[l-w, t, 3w, 3h]$, which is $3\times 3$ bigger than the original region and approximately covers the upper body of a person.

A Boosted Forest classifier is introduced after OHEM. Features from both the original and ``contextual'' regions are pooled using a fixed resolution of $5\times 5$, and then concatenated and input to a Boosted Forest classifier. We mainly follow~\cite{r18} to set the hyper-parameters of the BF classifier. Specifically, we bootstrap the training by six cascaded forests with an increasing number of trees: 64, 128, 256, 512, 1024 and 1536. The tree depth is set at 5. The initial training set contains all positive samples ($\sim$160k in the WIDER FACE training set) and randomly selected negative samples ($\sim$100k). After each stage, additional negative samples ($\sim$10k) are mined and added to the training set. At last, a forest of 2048 trees is trained as the final face detection classifier. Note that unlike an ordinary Boosted Forest, which equally initializes the confidence score of training samples, we directly use the ``faceness'' probability given by MP-RPN as the initial confidence score for each training sample.

\section{Experiments}
In this section, we first introduce the datasets used for training and evaluating our proposed face detector, and then compare the proposed MP-RCNN to state-of-the-art face detection methods on the WIDER FACE dataset~\cite{r04} and the FDDB dataset~\cite{r01}. The full implementation details of MP-RCNN used in the experiments are given in appendix A.

In addition, we conduct a set of detailed model analysis experiments to examine how each model component (e.g., detection branches, ``atrous'' convolution, OHEM, etc.) affects the overall detection performance. These can be found in appendix B. Moreover, the running time of our algorithm is reported in appendix C.

\subsection{Datasets}
WIDER FACE~\cite{r04} is a large public face detection benchmark dataset for training and evaluating face detection algorithms. It contains 32,203 images with 393,703 labeled human faces (each image has an average of 12 faces). Faces in this dataset have a high degree of variability in scale, pose, occlusion, lighting conditions, and image blur. Images in the WIDER FACE dataset are organized based on 61 event classes. For each event class, 40\%, 10\% and 50\% of the images are randomly selected for training, validation and test sets. Both the images and associated ground truth labels used for training and validation are available online\footnote{http://mmlab.ie.cuhk.edu.hk/projects/WIDERFace/index.html}. For the test set, only the images are available. The detection results must be submitted to an evaluation server administered by the authors of the WIDER FACE dataset in order to obtain Precision-Recall curves. Moreover, this test set was divided into three levels of difficulty by the authors of~\cite{r04} : ``Easy'', ``Medium'', ``Hard''. These categories were based on the detection rate of EdgeBox~\cite{r28}, so that the Precision-Recall curves need to be reported for each difficulty level\footnote{We have no knowledge about the difficulty level of the images in the test set. In fact, it is necessary to submit all predicted face boxes to the server, which then provided three ROC curves based on ``hard'', ``medium'' and ``easy'' partitions.}.

The other test set used in our experiments is the FDDB dataset~\cite{r01}, which is a standard database for evaluating face detection algorithms. It contains the annotations for 5,171 faces in a set of 2,845 images. Each image in FDDB dataset has less than two faces on average. These faces mostly have large sizes compared to those in the WIDER FACE dataset.

Our proposed MP-RCNN was trained on the training partition of the WIDER FACE dataset, and then evaluated on the WIDER FACE dataset test partition and the whole FDDB dataset. The validation partition of the WIDER FACE dataset is used in the model analysis experiments (appendix B) for comparing different model designs.

\subsection{Comparison to the state-of-the-art}
In this subsection, we compare the proposed MP-RCNN to state-of-the-art face detection methods on the WIDER FACE~\cite{r04} and FDDB datasets~\cite{r01}.

\textbf{\textit{Results on the WIDER FACE test set}} Here we compare the proposed MP-RCNN with all six strong face detection methods available on the WIDER FACE website: Two-stage CNN~\cite{r04}, Multiscale Cascade~\cite{r04}, Multitask Cascade~\cite{r29}, Faceness~\cite{r15}, Aggregate Channel Features (ACF) ~\cite{r12} and CMS-RCNN~\cite{r17}. Figure 4 shows the Precision-Recall curves and the Average Precision values of the different methods on the Hard, Medium and Easy partition of the WIDER FACE test set, respectively. On the hard partition, our MP-RCNN outperforms all six strong baselines by a large margin. Specifically, it achieves an increase of 9.6\% in Average Precision compared to the $2^{nd}$ place CMS-RCNN method. On the Easy and Medium partitions, our method both rank in $2^{nd}$ place, only lagging behind the recent CMS-RCNN method by a small margin. See Figure 6 in appendix D for some examples of the face detection results using the proposed MP-RCNN on the WIDER FACE test set.

\begin{figure*}
\begin{center}
\hfill
\subfigure[Hard Set]{\includegraphics[width=5.5cm]{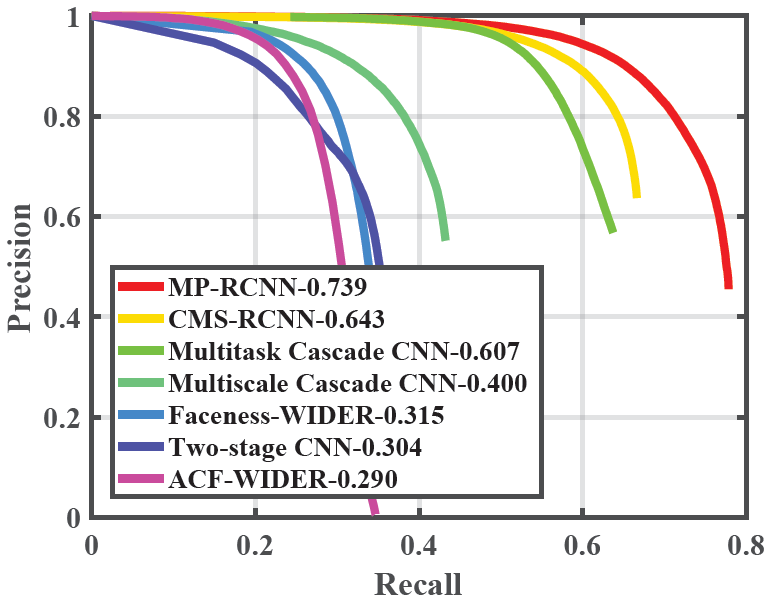}}
\hfill
\subfigure[Medium Set]{\includegraphics[width=5.5cm]{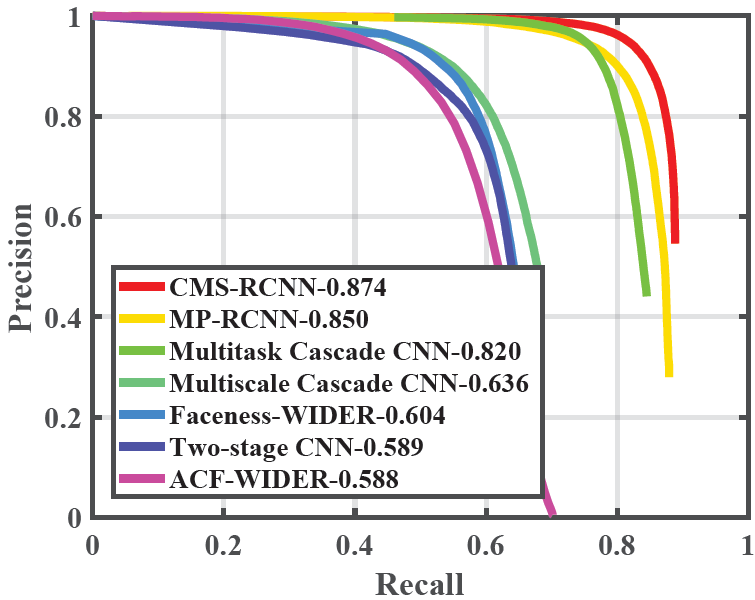}}
\hfill
\subfigure[Easy Set]{\includegraphics[width=5.5cm]{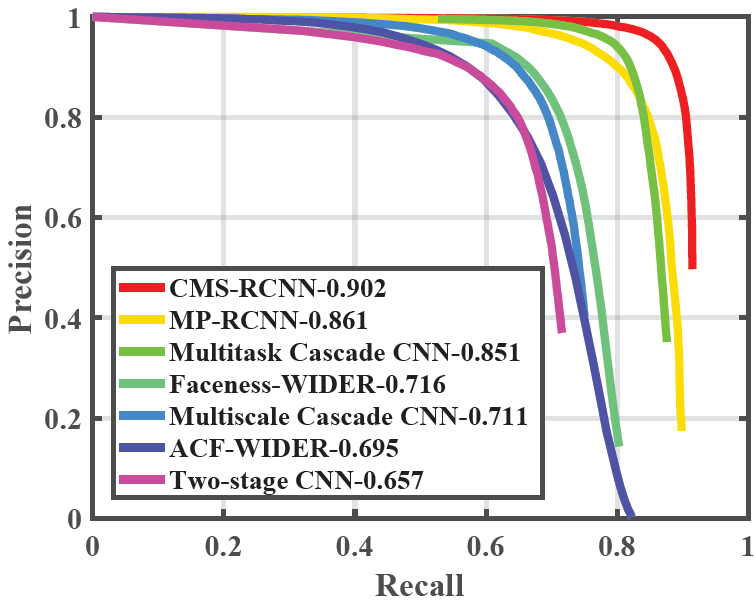}}
\hfill
\end{center}
\caption{Precision-Recall curves of our proposed MP-RCNN and other strong baselines, i.e., Two-stage CNN ~\cite{r04}, Multiscale Cascade~\cite{r04}, Multitask Cascade~\cite{r29}, Faceness~\cite{r15}, Aggregate Channel Features (ACF)~\cite{r12} and CMS-RCNN~\cite{r17}. Numbers in the legend show the average precision values. All methods follow Scenario-Int~\cite{r04} for training and testing, i.e., they are trained using the WIDER face training/validation sets, and tested on the WIDER FACE test partition.}
\end{figure*}

\begin{figure}[t]
\begin{center}
   \includegraphics[width=0.98\linewidth]{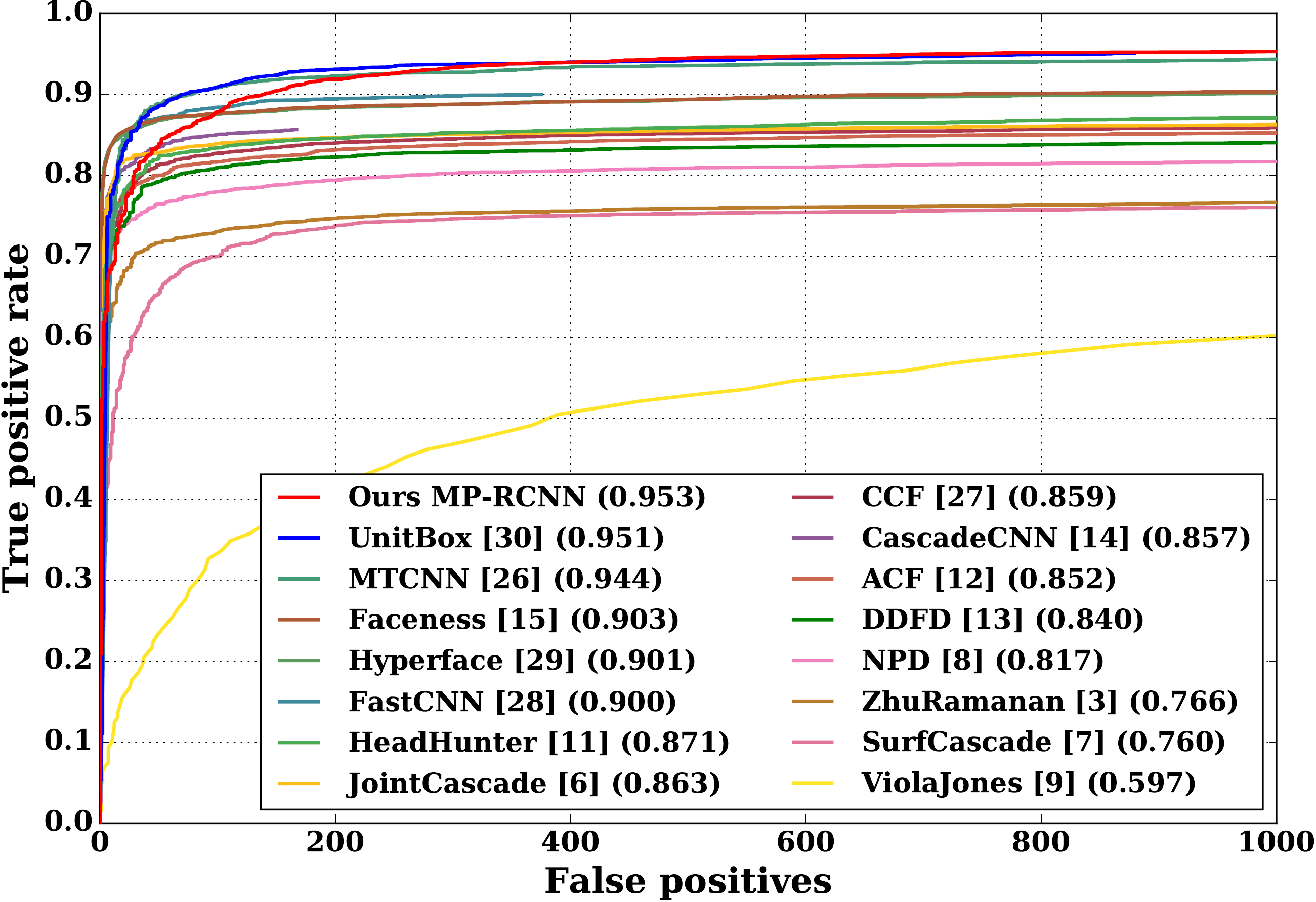}
\end{center}
   \caption{ROC curves of the proposed MP-RCNN and other published strong methods on the FDDB dataset ~\cite{r02}. Numbers in the legend show the average recall rates.}
\label{fig5_onecol}
\end{figure}

\textbf{\textit{Results on the FDDB dataset}} To show the general face detection capability of the proposed MP-RCNN method, we directly apply the MP-RCNN previously trained on the WIDER FACE training set to the FDDB dataset. We also make a comprehensive comparison with 15 other typical baselines: ViolaJones~\cite{r09}, SurfCascade~\cite{r07}, ZhuRamanan~\cite{r03}, NPD~\cite{r08}, DDFD~\cite{r13}, ACF~\cite{r12}, CascadeCNN~\cite{r14}, CCF~\cite{r30}, JointCascade~\cite{r06}, HeadHunter~\cite{r11}, FastCNN~\cite{r31}, Faceness~\cite{r15}, HyperFace~\cite{r32}, MTCNN~\cite{r29} and UnitBox~\cite{r33}. The evaluation is based on a discrete score criterion, that is, if the ratio of the intersection of a detected region with an annotated face region is greater than 0.5, a score of 1 is assigned to the detected region, and 0 otherwise. As shown in Figure~\ref{fig5_onecol}, the proposed MP-RCNN outperforms ALL of the other 15 methods and has the highest average recall rate (0.953). See Figure~\ref{fig7_onecol} in appendix E for some examples of the face detection results on the FDDB dataset.

\section{Conclusion}
We have proposed MP-RCNN, an accurate face detection method for tackling the challenge of large-scale variation in unconstrained face detection. Most previous methods extract the same features for faces at different scales. This neglects the face pattern variations due to scale changes and thus fails to detect both large and tiny faces with high accuracy. In this paper, we introduce MP-RCNN, which utilizes a newly proposed Multi-Path Region Proposal Network (MP-RPN) to extract features at various intermediate network layers. These features possess different receptive field sizes that approximately match the facial patterns at three different scales. This leads to high detection accuracy for faces across a large range (from $8\times 8$ to $900 \times 900$) of facial scales. 

MP-RCNN also employs a boosted forest classifier as the second stage, which uses the deep features pooled from MP-RPN to further boost face detection performance. We observe that although MP-RCNN is designed mainly to deal with the challenge of scale variation, the powerful feature representation of deep networks also enables a high level of robustness to variations in pose, occlusion, illumination, out-of-focus blur and background clutter. Experimental results demonstrate that our proposed MP-RCNN consistently achieves the best performance on both the WIDER FACE and FDDB datasets. In the future, we intend to leverage this across-scale detection ability to other tiny object detection tasks, e.g., facial landmark localization of small faces.


\section*{Acknowledgments}
The authors would like to acknowledge the financial support of the Natural Sciences and Engineering Research Council of Canada (NSERC) and the McGill Engineering Doctoral Award (MEDA). They would also like to thank the support of the NVIDIA Corporation for the donation of a TITAN X GPU through their academic GPU grants program.



\newcommand{\BIBdecl}{\setlength{\itemsep}{0.25 em}}
\bibliographystyle{IEEEtran}
\bibliography{IEEEabrv,paper_abridged}
%



\newpage
\section*{Appendix}

\subsection{Implementation Details}
The code of MP-RPN and the deep feature extraction was built using Caffe~\cite{r25}, and the Boosted Forest was based on Piotr\textquotesingle s Computer Vision Matlab Toolbox~\cite{r26}.

Before training and testing, each full image of arbitrary size was resized such that its shorter edge had $N$ pixels ($N=900$ in the WIDER FACE dataset and $400$ in the FDDB dataset). 

For MP-RPN training, an anchor was assigned as a positive sample if it had an Intersection-over-Union (IOU) ratio greater than 0.5 with any ground truth box, and as a negative sample if it had an IOU ratio less than 0.3 with any ground truth box. Each mini-batch contains 1 image and 768 sampled (using OHEM) anchors, 256 for each detection branch. The ratio of positive and negative samples is 1:3 for all detection branches. The CNN backbone (from Conv1\_1 to Conv5\_3 in Figure~\ref{fig2}) was a truncated VGG-16 net~\cite{r19} pre-trained on the ImageNet dataset~\cite{r27}. The weights of all the other convolutional layers were randomly initialized from a zero-mean Gaussian distribution with standard deviation 0.01. We fine-tuned the layers from conv3\_1 and up, using a learning rate of 0.0005 for 80k mini-batches, and 0.0001 for another 40k mini-batches on the WIDER FACE training dataset. A momentum of 0.9 and a weight decay of 0.0005 were used.
Face proposals produced by MP-RPN are post-processed individually for each detection branch in the following way. First, non-maximum suppression (NMS) with a threshold of 0.7 was adopted to filter face proposals based on their classification scores. Then the remaining face proposals were ranked by their scores. For BF training, 150, 40, 10 top-ranked proposals in an image were selected from Det-4, Det-16 and Det-32, respectively. At test time, the same number (150, 40, 10) of proposals were selected from the corresponding branch, and finally all output proposals from the different branches were merged by NMS with a threshold of 0.5.

\subsection{Model Analysis}
In this subsection, we discuss controlled experiments on the validation set of the WIDER FACE dataset to examine how each model component affects the overall detection performance. Note that in order to save training time, experiment 1-3 employed face detection models trained for 30k iterations on only 11 out of the total 61 event classes. The learning rate was selected to be 0.0005 for the first 20k iterations, and 0.00005 for the remaining 10k iterations. Other hyper-parameters were determined as stated earlier in appendix A. The selected event classes are the first eleven classes (i.e., Traffic, Parade, Ceremony, People Marching, Concerts, Award Ceremony, Stock Market, Group, Interview, Handshaking and Meeting), which take up about 1/5 of the whole training set. In Experiment 4, the face detection model was trained with the whole WIDER FACE training set (61 event classes). All hyper-parameters in Experiment 4 were the same as stated in appendix A.

\textbf{\textit{Experiment-1: The roles of individual detection layers}}
Table \ref{table1} shows the detection recall rates of the various detection branches as a function of face height in pixels. We observe that each detection branch has the highest detection recall for the faces that match its scale. The combination of all detection branches (the last row of Table \ref{table1}) achieves the highest recall for faces of all scales. Note that the recall rate for small scale faces (8$\le$height$\le$32) is much lower than that of medium scale faces (32$<$height$\le$360) and large scale faces (360$<$height$\le$900), indicating the obvious expectation of the increasing difficulty of face detection as scale drops.

\begin{table}[h!]
\begin{center}
\caption{Detection recall of various detection branches on WIDER FACE validation set as a function of face height in pixels}
\label{table1}
    \begin{tabular}{  p{1cm} | p{1.5cm} | p{1.5cm} | p{1.5cm}| p{1cm}}
    \hline
    Branch & 8$\le$height\newline$\le$32 & 32$<$height\newline$\le$360 & 360$<$height\newline$\le$900 & All scales \\ \hline
    Det-4 & \textbf{0.7994} & 0.4173 & 0 & 0.6683 \\ 
    Det-16 & 0.2862 & \textbf{0.9076} & 0 & 0.9759 \\ 
    Det-32 & 0 & 0.0488 & \textbf{0.9919} & 0.02 \\ 
    Combined & \textbf{0.8035} & \textbf{0.9263} & \textbf{0.9919} & 0.8454 \\
    \hline
    \end{tabular}
\end{center}
\end{table}

\textbf{\textit{Experiment-2: The roles of atrous convolutional layers}}
Table \ref{table2} shows the detection recall rates of the proposed MP-RPN in terms of different design options (with/without ``atrous'' convolution and with/without OHEM). By comparing rows 1 and 3, as well as 2 and 4, we observe that the inclusion of the ``atrous'' convolution trick increases the detection recall rate of all branches.

\begin{table}[h!]
\begin{center}
\caption{Detection Recall of MP-RPN with different options on the WIDER FACE validation set as a function of face height in pixels}
\label{table2}
    \begin{tabular}{  p{0.8cm} | p{0.8cm} | p{1.2cm} | p{1.2cm} | p{1.2cm}| p{1cm}}
    \hline
    \multicolumn{2}{c|}{MP-RPN option} & \multirow{2}{3cm}{8$\le$height\newline$\le$32} & \multirow{2}{3cm}{32$<$height\newline$\le$360} & \multirow{2}{3cm}{360$<$height\newline$\le$900} & \multirow{2}{*}{All scales}\\
    \cline{1-2}
    Atrous? & OHEM? & & & & \\  \hline
    $\times$ & $\times$ & 0.7524 & 0.9196 & 0.9839 & 0.8083 \\ 
    $\times$ & \checkmark & 0.7813 & 0.9059 & 0.9839 & 0.8239 \\ 
    \checkmark & $\times$ & 0.8031 & 0.9214 & 0.9919 & 0.8435 \\ 
    \checkmark &\checkmark & \textbf{0.8035} & \textbf{0.9263} & \textbf{0.9919} & \textbf{0.8454} \\
    \hline
    \end{tabular}
\end{center}
\end{table}

\textbf{\textit{Experiment-3: The roles of the OHEM layers}} By comparing rows 1 and 2, as well as 3 and 4 in Table \ref{table2}, we can conclude that, in most cases, the inclusion of the OHEM layer increases the detection recall rate. However, in the absence of ``atrous'' convolution, the use of OHEM layer causes a slight recall drop for medium size faces (32$<$height$\le$360). By comparing rows 1 and 4, we see observe that the simultaneous inclusion of ``atrous'' convolution and OHEM consistently increases the detection recall of all face scales.

\begin{figure*}
\begin{center}
\includegraphics[width=0.9\linewidth]{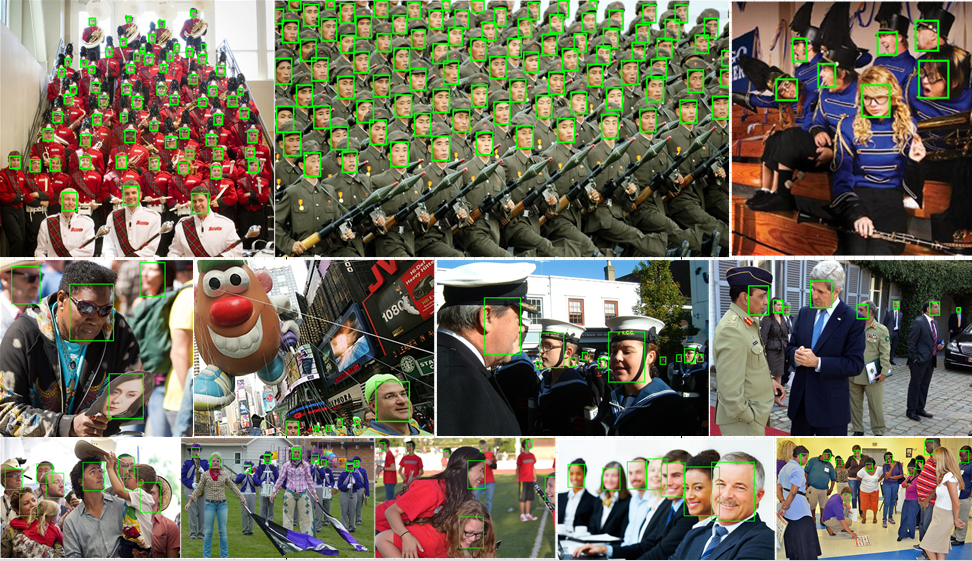}
\end{center}
   \caption{Example of face detection results on the WIDER FACE test set~\cite{r01} using the proposed MP-RCNN method.}
\label{fig6}
\end{figure*}

\textbf{\textit{Experiment-4: The roles of BF with various options}} Table \ref{table3} displays the average precision of various Boosted Forest (BF) options. We observe that although MP-RPN already achieves high average precision as a stand-alone face detector, the inclusion of a BF classifier further boosts the detection performance for faces of all levels of difficulty. Specifically, a BF classifier with ``face'' features (features pooled from the original proposal regions\footnote{See Section 3.B for details.}) achieves a relatively higher average precision gain for ``easy'' and ``medium'' faces, but a lower average precision gain for ``hard'' faces, compared to a BF classifier with ``context'' features (features pooled from a larger region surrounding the original proposal regions\footnote{See Section 3.B for details.}). When pooling complementary ``face'' and ``context'' features, the BF classifier achieves the highest gain for all ``Easy'', ``Medium'' and ``Hard'' faces.

\begin{table}[h!]
\begin{center}
\caption{Average Precision of Boosted Forest (BF) Classifier with various options on WIDER FACE validation set.}
\label{table3}
    \begin{tabular}{ p{3.6cm} | p{1cm} | p{1cm} | p{1cm}}
    \hline
    \multirow{2}{*}{Method} & \multicolumn{3}{|c}{Average Precision} \\ 
    \cline{2-4}
    & Easy & Medium & Hard \\ \hline
    MP-RPN & 0.856 & 0.848 & 0.722 \\ 
    MP-RPN + BF(face) & 0.860 & 0.851 & 0.726 \\ 
    MP-RPN + BF(context)& 0.857 & 0.849 & 0.728 \\ 
    MP-RPN + BF(face+context)& \textbf{0.862} & \textbf{0.852} & \textbf{0.734} \\ \hline
    \end{tabular}
\end{center}
\end{table}

\subsection{Average processing time}
We randomly selected 100 images from the WIDER FACE validation set. An image patch of resolution $640 \times 480$ was cropped from the center of each image\footnote{If the original image had a height less than 640 or a width less than 480 pixels, we padded the cropped image patch from the bottom and the right with zeros to make it exactly $640 \times 480$.}, thus creating 100 new images. Both the proposed MP-RCNN and the classical Viola-Jones algorithm~\cite{r09} were employed to process these 100 images. The average processing time per image is shown in Table \ref{table4} below. Note that in order to guarantee a fair comparison, both algorithms were tested on a 3.5 GHz 8-core Intel Xeon E5-1620 server with 64GB of RAM, and the image loading time was excluded from the processing time for both algorithms. The Viola-Jones algorithm\footnote{We used the code provided by the OpenCV website: \url{http://docs.opencv.org/2.4/modules/objdetect/doc/cascade_classification.html}. The face model used in the code was ``haarcascade\_frontalface\_default''.} used only CPU resources. An Nvidia GeForce GTX Titan X GPU was used for the CNN computations in MP-RCNN.

\begin{table}[h!]
\begin{center}
\caption{A comparison of average processing time}
\label{table4}
    \begin{tabular}{  p{1.8cm} | p{2.25cm} | p{2.25cm}}
    \hline
    Method & Programming Language & Average processing time (sec.) \\ \hline
    Viola-Jones & C++ & 0.092 \\
    MP-RCNN & Matlab and C++ & 0.216 \\
    \hline
    \end{tabular}
\end{center}
\end{table}

From Table \ref{table4}, we observe that the proposed MP-RCNN runs at about 4.6 FPS compared to the 10.9 FPS obtained by classical Viola-Jones algorithm.

\subsection{Face detection results on WIDER FACE test set}
Figure~\ref{fig6} shows some examples of the face detection results using the proposed MP-RCNN on the WIDER FACE test set.

\subsection{Face detection results on FDDB}
\begin{figure}[t]
\begin{center}
   \includegraphics[width=0.9\linewidth]{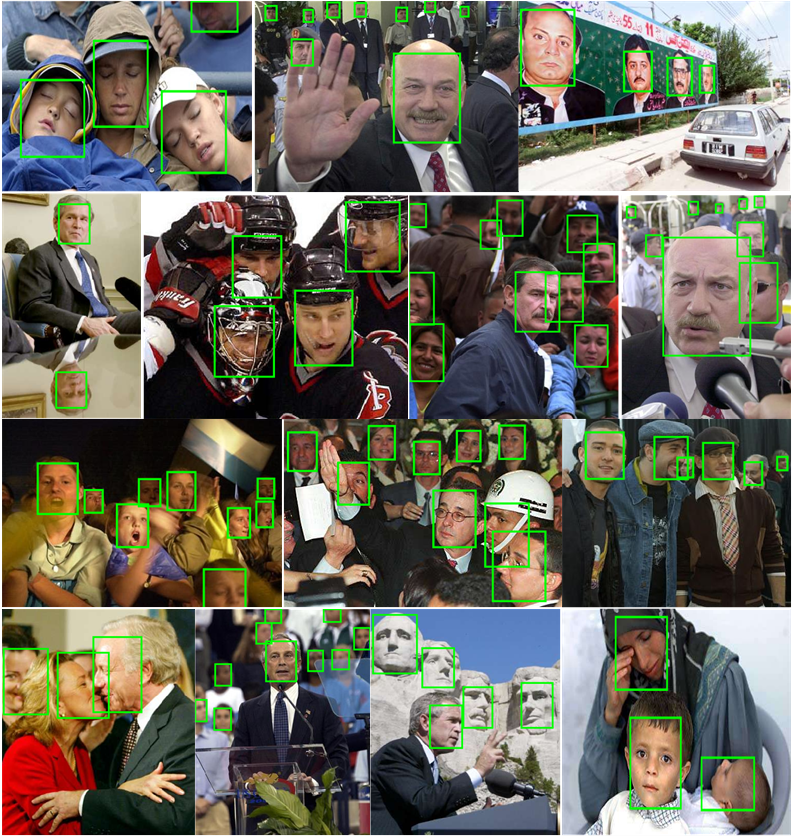}
\end{center}
   \caption{Example of face detection results on the FDDB dataset~\cite{r02} using the proposed MP-RCNN method.}
\label{fig7_onecol}
\end{figure}

Figure~\ref{fig7_onecol} shows some examples of the face detection results using the proposed MP-RCNN on FDDB dataset.

\end{document}